%% file: paper.tex
\pdfoutput=1

\documentclass[11pt]{article}

\usepackage[]{acl}

\usepackage{times}
\usepackage{latexsym}

\usepackage[T1]{fontenc}

\usepackage[utf8]{inputenc}

\usepackage{microtype}

\usepackage{inconsolata}

\usepackage{amsmath,amssymb,amsfonts}
\usepackage{url,enumitem}
\usepackage{booktabs}
\usepackage{xspace}
\usepackage{graphicx}
\usepackage{subcaption}
\usepackage{comment}
\usepackage[normalem]{ulem}
\usepackage{makecell}
\usepackage{multirow}
\usepackage{diagbox}

\usepackage{mathtools}
\usepackage{svg}
\usepackage{algorithm,algorithmicx,algpseudocode}
\usepackage{listings}
\newcommand{\Sref}[1]{\S\ref{#1}}

\DeclareMathOperator{\argmax}{argmax}

\title{Understanding In-Context Learning via Supportive Pretraining Data
}

\author{
  Xiaochuang Han$^{\spadesuit\clubsuit}$\thanks{\ \*  Work done during an internship at Meta AI.} 
  \quad D\'aniel Simig$^\clubsuit$ \quad Todor Mihaylov$^\clubsuit$ \\ \quad \textbf{Yulia Tsvetkov}$^\spadesuit$ \quad \textbf{Asli Celikyilmaz}$^\clubsuit$ \quad \textbf{Tianlu Wang}$^\clubsuit$\\
  $^\clubsuit$Meta AI \\
  $^\spadesuit$University of Washington \\
        {\tt \small \{xhan77, yuliats\}@cs.washington.edu \quad simigd@gmail.com \quad \{tbmihaylov, aslic, tianluwang\}@meta.com}
}

\begin{document}
\maketitle

\begin{abstract}
In-context learning (ICL) improves language models' performance on a variety of NLP tasks by simply demonstrating a handful of examples at inference time. It is not well understood why ICL ability emerges, as the model has never been specifically trained on such demonstrations. Unlike prior work that explores implicit mechanisms behind ICL, we study ICL via investigating the \emph{pretraining data}. Specifically, we first adapt an iterative, gradient-based approach to find a small subset of pretraining data that \emph{supports} ICL. We observe that a continued pretraining on this small subset significantly improves the model's ICL ability, by up to 18\%. We then compare the supportive subset constrastively with random subsets of pretraining data and discover: (1) The supportive pretraining data to ICL do \emph{not} have a higher domain relevance to downstream tasks. (2) The supportive pretraining data have a higher mass of rarely occurring, long-tail tokens. (3) The supportive pretraining data are \emph{challenging} examples where the information gain from long-range context is below average, indicating learning to incorporate difficult long-range context encourages ICL. Our work takes a first step towards understanding ICL via analyzing instance-level pretraining data. Our insights have a potential to enhance the ICL ability of language models by actively guiding the construction of pretraining data in the future.

\end{abstract}

\section{Introduction}
\label{sec:introduction}
\input{sections/intro}

\section{Finding supportive pretraining data for in-context learning}
\label{sec:method}

\input{sections/method.tex}

\section{Analyzing supportive pretraining data for in-context learning}
\label{sec:analysis}
\input{sections/analysis.tex}

\section{Related Work}
\label{sec:related_work}
\input{sections/related_work.tex}

\section{Conclusion}
\input{sections/conclusion.tex}

\section*{Limitations}
\input{sections/limitation.tex}

\section*{Acknowledgements}
We thank Naman Goyal, Anjali Sridhar, Zeyu Liu, Victoria Lin, Mengzhou Xia, Weijia Shi, Jiacheng Liu, Hao Zhu, and Tianxing He for helpful discussions. 
We also thank the anonymous ACL reviewers and all members of TsvetShop for the valuable feedback. 
This research is supported in part by the Office of the Director of National Intelligence (ODNI), Intelligence Advanced Research Projects Activity (IARPA), via the HIATUS Program contract \#2022-22072200004. The views and conclusions contained herein are those of the authors and should not be interpreted as necessarily representing the official policies, either expressed or implied, of ODNI, IARPA, or the U.S. Government. The U.S. Government is authorized to reproduce and distribute reprints for governmental purposes notwithstanding any copyright annotation therein.

\bibliography{my_cites}
\bibliographystyle{acl_natbib}

\appendix

\input{sections/appendix.tex}

\end{document}

%% file: sections/intro.tex
\begin{figure}[t]
    \centering
    \includegraphics[width=\linewidth]{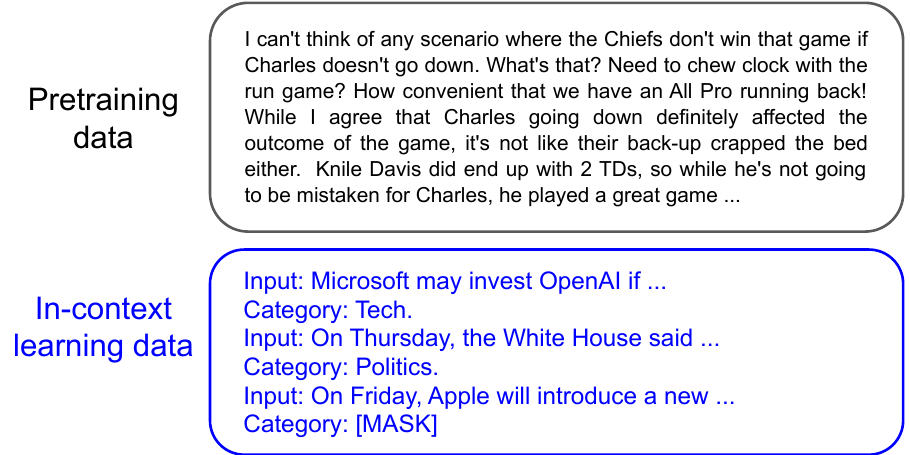}
    \caption{An example from the pretraining data of OPT \citep{zhang2022opt} and an illustrative in-context learning example of topic classification. 
    The in-context learning task data can be drastically different from pretraining instances, both in content and format.}
    \label{fig:icl_first_fig}

\end{figure}


In-context learning in NLP has drawn tremendous attention recently \citep{dong2022survey}. Unlike traditional learning paradigms that rely on training or finetuning models, 
in-context learning only provides a handful of demonstration examples to language models as a prefix to the test input, without any parameter updates. 
In-context learning has shown superior performance on a range of NLP tasks~\cite{NEURIPS2020_1457c0d6, zhang2022opt,chowdhery2022palm, hoffmann2022training}, 
but the origin and reason of this emergent ability remain under-investigated. 
In-context learning is surprising since language models have not been explicitly trained to learn from demonstration examples \citep{xie2022an}. As shown in an illustrative scenario in \autoref{fig:icl_first_fig}, a typical pretraining data instance is highly different from an in-context learning example for downstream tasks, in both content and format.


Prior work have attempted to answer \emph{what} in-context learning is, through empirically investigating useful and irrelevant attributes of the demonstration examples \citep{min2022rethinking, zhang2022robustness}, or theoretically proving certain synthetic language models implicitly do Bayesian inference with demonstrations \citep{xie2022an}. Furthermore, recent work have drawn connections between the mechanism of in-context learning and standard learning algorithms, such as regression, nearest neighbor, and gradient descent \citep{olsson2022context, akyurek2022learning, dai2022can, von2022transformers}. 

Differently, in this work we are interested in understanding \emph{from where} the in-context learning ability is acquired, through a perspective of pretraining data. Although not many, some recent work have investigated this direction. For instance,
\citet{shin2022effect} pretrain a variety of language models on different corpora. They study correlations between attributes of pretraining datasets and in-context learning performance, 
at a relatively coarse dataset-level. 
\citet{chan_data_2022} construct pretraining data with different attributes and discover that some distributional properties of the data drive the emergence of in-context learning. However, their experiment is limited to synthetic data of image-label pairs.



In this work, we investigate a large language model OPT \citep{zhang2022opt} and its pretraining data. 
We first hypothesize that there exists some specific pretraining data instances that are particularly helpful to the model's in-context learning ability. 
As an attempt to find such instances, we adapt an iterative, gradient-based method ORCA \citep{han2022orca} to search within OPT's pretraining corpus. The process is guided by the gradients of the in-context learning data from downstream tasks, and we refer to the identified subset as supportive pretraining data to in-context learning following~\citet{han2022orca}. Furthermore, we quantitatively verify through a perturbative continued pretraining, that the supportive subset does improve the model's in-context learning performance on downstream tasks, while not affecting a spurious zero-shot performance (\Sref{sec:method}).

We then analyze the identified supportive data in contrast to the general pretraining data, to obtain data features particularly relevant to in-context learning. 
We specifically approach from three aspects: the domain relevance to downstream tasks, the token frequency distribution, and the information gain of incorporating long-range pretraining context. Our major findings include: 
\textbf{(1)} Compared to general pretraining data, the supportive data do \emph{not} have a higher domain relevance to the downstream tasks. 
\textbf{(2)} The supportive pretraining data contain a relatively higher amount of rarely occurring, long-tail tokens. 
\textbf{(3)} The supportive pretraining data are \emph{challenging} examples in incorporating long-range context for language modeling (\Sref{sec:analysis}).

Our work offers a first step towards interpreting in-context learning in NLP tasks via analyzing instance-level pretraining data. 
We believe it can help improve the transparency and interpretability of language models' in-context learning behavior. 
Our analysis can also pave the way to improved in-context learning in the future by informing pre-training data construction. 


%% file: sections/method.tex
\label{sec:method}

\citet{han2022orca} propose an iterative, gradient-based method ORCA to find supportive pretraining data of BERT \citep{Devlin2019BERTPO} under a vanilla zero-shot prompting setup. 
In this section, we provide some background and adapt ORCA for large language models in a setting of in-context learning (ICL), finding supportive pretraining data for downstream tasks with demonstration examples.\footnote{Identifying important training data for an inference time model output is an estabilished topic in model interpretability, with various prior work measuring data importance via variants of gradient similarity \citep{Koh2017UnderstandingBP,Pruthi2020EstimatingTD}. 
However, these methods are prohibitively expensive to be applied to large-scale pretraining data. 
Concurrent to our work, \citet{Guu2023SimfluenceMT} propose an interesting method to model the importance of individual training examples by simulating training runs, but it is also on a scale of finetuning instead of pretraining.}


\subsection{Methodology}
Assume we have a pretrained language model (LM) $\theta$ and data pairs $(\boldsymbol{x}, \boldsymbol{y})$ representing the inputs and ground truth outputs of task $D_{\text{task}}$. Both $\boldsymbol{x}$ and $\boldsymbol{y}$ are in natural language. For classification tasks, the target labels can be converted to natural language via verbalizers \citep{Schick2021ExploitingCF}. 

\paragraph{Zero-shot prompting} 
A pretrained language model can be applied to perform downstream tasks via zero-shot prompting \citep[e.g.,][]{Petroni2019LanguageMA}. For classification tasks, the language model $\theta$ outputs a candidate answer with top probability, $\argmax_{\boldsymbol{y}' \in \mathcal{Y}} p_\theta(\boldsymbol{y}' \mid \boldsymbol{x}) = \argmax_{\boldsymbol{y}' \in \mathcal{Y}} \prod_{t=0}^{t<|\boldsymbol{y}'|} p_\theta(y'_t \mid \boldsymbol{x}, \boldsymbol{y}'_{<t})$, where $\mathcal{Y}$ contains all candidate answers $\boldsymbol{y}'$. 
For generation tasks, outputs can be obtained by sampling autoregressively from $\theta$ conditioned on $\boldsymbol{x}$ \citep[e.g.,][]{holtzman2019curious}. 
This is a zero-shot scenario with no demonstration examples.

\paragraph{In-context learning}
Instead of modeling $p_\theta(\boldsymbol{y} \mid \boldsymbol{x})$, ICL estimates $p_\theta(\boldsymbol{y}\mid\{ (\boldsymbol{x}_\text{demo}, \boldsymbol{y}_\text{demo}) \}, \boldsymbol{x})$, prepending the original model input with several demonstration examples $(\boldsymbol{x}_\text{demo}, \boldsymbol{y}_\text{demo})$ sampled from the target task $D_{\text{task}}$. The language model $\theta$ is never trained on the task data with demonstrations. However, we can form a loss on the in-context data as a surrogate for $\theta$'s ICL performance, which will be used for a later guidance step, $L_\theta^{\text{ICL}}(\boldsymbol{x}, \boldsymbol{y}) = -\log p_\theta(\boldsymbol{y} \mid \{ (\boldsymbol{x}_\text{demo}, \boldsymbol{y}_\text{demo}) \}, \boldsymbol{x}) = -\log \prod_{t=0}^{t<|\boldsymbol{y}|} p_\theta(y_t \mid \{ (\boldsymbol{x}_\text{demo}, \boldsymbol{y}_\text{demo}) \}, \boldsymbol{x}, \boldsymbol{y}_{<t})$. 

\paragraph{Pretraining}
The pretraining data of $\theta$ often consists of texts $\boldsymbol{w}$ from large, general-domain corpora. During pretraining, the LM $\theta$ is updated via stochastic gradient descent with a loss to reconstruct $\boldsymbol{w}$ given a prefixing context, $L_\theta^{\text{PT}}(\boldsymbol{w}) = -\log \prod_{t=0}^{t<|\boldsymbol{w}|} p_\theta(w_t \mid \boldsymbol{w}_{<t})$. 

\paragraph{Supportive pretraining data}
Our goal is to locate what pretraining data $\boldsymbol{w}$ if upweighted would be most helpful to the LM $\theta$'s ICL ability. Following ORCA \citep{han2022orca}, we use the similarity between gradients $\nabla_\theta L_\theta^{\text{PT}}(\boldsymbol{w})$ and $\nabla_\theta L_\theta^{\text{ICL}}(\boldsymbol{x},\boldsymbol{y})$ iteratively to find such supportive pretraining data. 
We show details of our adapted algorithm ORCA-ICL in \autoref{fig:algo}. 
The algorithm finds pretraining data that exert a gradient to $\theta$ similarly as a group of guidance ICL task data would. 
$\nabla_\theta L_\theta^{\text{ICL}}(\boldsymbol{x},\boldsymbol{y})$ provides a guidance for the direction the model parameters \emph{should} be updated towards to be better at ICL, while 
$\nabla_\theta L_\theta^{\text{PT}}(\boldsymbol{w})$ approximates how the direction the model parameters \emph{would} be updated based on individual pretraining instances. 
We conduct a multi-iteration process (a total of $M$ iterations each selecting $k$ supportive instances) to mitigate noise.\footnote{Additionaly according to \citet{han2022orca}, this may prevent selecting examples associated with only one class of the task, a case of poor calibration.} SGD denotes an one-pass stochastic gradient descent to mimick an incremental upweight to the selected data, with a minimum number of steps to prevent overfitting. The resulting supportive set $S$ has a very small size (under 2000 in this work).\footnote{More details of the ORCA algorithm can be found in \citet{han2022orca}.} 

\paragraph{Verifying supportiveness}
To quantitatively evaluate the supportiveness of the selected set of pretraining data, we perform an one-pass gradient descent on the original LM with the selected set $S$, which mimics a \emph{perturbative continued pretraining} with a minimum number of updates: $\theta_M \leftarrow \underset{S}{\text{SGD}}(\theta_0)$. 
We then benchmark this perturbed model ($\theta_M$) with the original model ($\theta_0$) and a model perturbed with a random set of pretraining data. We expect the perturbed model using our selected supportive pretraining data to achieve a better ICL performance. 

\algrenewcommand\algorithmicindent{0.4em}%
\begin{figure}[t]
\begin{minipage}[t]{0.48\textwidth}
\begin{algorithm}[H]
  \caption{ORCA-ICL} \label{alg:training}
  \small
  \begin{algorithmic}[1]
    \State Load a pretrained language model as $\theta_0$
    \For{$i \leftarrow 1, M$}
      \If{$i=1$}
      \State $S_1 \leftarrow \underset{\boldsymbol{w} \in D_{\text{PT}}}{\mathrm{argtop\text{-}}k} [ \cos(\nabla_\theta L_{\theta_0}^{\text{PT}}(\boldsymbol{w}), \nabla_\theta \underset{ D_{\text{task}}}{\sum} L_{\theta_0}^{\text{ICL}}(\boldsymbol{x},\boldsymbol{y})) ]$
      \State $\theta_1 \leftarrow \underset{S_1}{\text{SGD}}(\theta_0)$
      \Else 
      \State $S_i \leftarrow \underset{\boldsymbol{w} \in D_{\text{PT}}}{\mathrm{argtop\text{-}}k} [ \cos(\nabla_\theta L_{\theta_0}^{\text{PT}}(\boldsymbol{w}), \nabla_\theta \underset{ D_{\text{task}}}{\sum} L_{\theta_{i-1}}^{\text{ICL}}(\boldsymbol{x},\boldsymbol{y})) ]$
      \State $\theta_i \leftarrow \underset{\cup_{j=1}^{i} S_j}{\text{SGD}}(\theta_0)$
      \EndIf
      \EndFor
      \State Return supportive pretraining data $S \leftarrow \cup_{i=1}^{M} S_i$
      
  \end{algorithmic}
\end{algorithm}
\end{minipage}

\caption{ORCA-ICL, an iterative gradient-based selection of supportive pretraining data for ICL.}
\label{fig:algo}
\end{figure}

\subsection{Setup}
\paragraph{Language model}
Throughout the work, we use a pretrained, autoregressive OPT-6.7B \citep{zhang2022opt} as our LM $\theta$. 

\paragraph{Tasks}
In this work, we focus on classification problems and first retrieve 48 classification-based tasks from Natural Instructions v2 \citep[NI-v2,][]{wang2022benchmarking}. We apply the LM on the tasks with both a zero-shot and in-context learning setup. We extract tasks that achieve at least 10\% better performance with in-context demonstrations. We group 17 tasks that satisfies the constraint and further select 6 typical tasks among them: 

\textbf{SST-2}: Movie review sentiment classification \citep{socher2013recursive}. 
\textbf{AG News}: News topic classification \citep{Zhang2015CharacterlevelCN}. 
\textbf{Story Cloze Test}: Story coherence classification \citep{mostafazadeh2017lsdsem}. 
\textbf{SMS Spam Collection}: Spam classification \citep{almeida2011contributions}. 
\textbf{Sentiment 140}: Tweet sentiment classification \citep{go2009twitter}. 
\textbf{TweetQA}: Answer verification \citep{xiong2019tweetqa}. 

For each task, we randomly sample 500 examples with a balanced class distribution 
as $D_{\text{task}}$, guiding the ORCA-ICL algorithm. The quantitative evaluation is performed on the full dataset. 
For ICL, for each instance in the task data, we randomly sample 4 demonstration examples under each candidate class defined in the task.\footnote{The sampling of demonstration examples is independent across test instances to mitigate potential spurious correlations.} 
The order of demonstration examples in the context is randomly shuffled. 
The template and verbalizer of each task follows the original NI-v2 dataset, though we did not include the task instructions, as the focus of this work is in-context learning with demonstration examples. 


\paragraph{Pretraining} 
Considering the size of pretraining data $D_{\text{PT}}$, we include an as large portion of OPT's pretraining data as possible under a reasonable budget. Specifically, in this work we use a total of 2.5M pretraining instances each consists of 2048 tokens.\footnote{The total 5B tokens are about 3\% of OPT's 180B full pretraining data.} 
For computing efficiency, we use intra-layer model parallelism \citep{shoeybi2019megatron} and fully sharded data parallel \citep{FSDP}.\footnote{This groups 4 input data for each backward pass in our setup. The 4 instances receive a same gradient similarity score, equivalent to an aggregated instance 4 times of the length.} 

\paragraph{Implementation Details}
We run ORCA-ICL with a maximum of $M=5$ iterations. In each iteration we extract $k=400$ pretraining instances with top gradient similarity with the ICL task data. 
We use a batch size of 16 and learning rate of 2e-5 for the one-pass gradient descent with an Adam optimizer \citep{kingma2014adam}. This results in a total of 125 updates\footnote{The one-pass descent has $\frac{M * k}{\text{batch size}}$ steps.} to the original LM after all iterations as the perturbative continued pretraining. 

\subsection{Results
} 

\paragraph{Perturbative continued pretraining} 
As the main evaluation of the supportive pretraining data obtained by ORCA-ICL, we perform perturbative continued pretraining on both the selected supportive data and random pretraining data as a control. \autoref{tab:main_result} shows the main results of task accuracy. The leftmost column shows a source task $D_{\text{task}}$ guiding the selection of supportive pretraining data. At each row, we evaluate the perturbed model ($\underset{S}{\text{SGD}}(\theta_0)$) on all 6 tasks. 
The ICL performance of the original LM is reported in the headers of the table. 

In each cell of the table, the top number shows the continued pretraining result with the supportive data we identified. We consider $M \in [1, 5]$ iterations as a hyperparameter and report result with a best $M$. 
We want to know \emph{at a same size of selection}, how our identified subset performs compared to random pretraining data. 
We therefore run random selection with 5 seeds, and the bottom number of the cell shows the continued pretraining result with random data at a same size of our selection, accompanied by a standard deviation. 
The performance of our selection is bolded when the performance difference with random selection exceeds one standard deviation. 

The diagonal cells show the performance of perturbed models on the same task used for selecting supportive data. We observe on 4 of the 6 source tasks, our selection of supportive pretraining data is effective. For the cross-task performance, we observe on 5 of the 6 source tasks, our selection is effective for at least three tasks.\footnote{Negative result is observed with TweetQA, on which we conjecture the patterns in the demonstration examples are more difficult to transfer to the test input (e.g., factual knowledge instead of sentiment indicators).}  
We conclude that \textbf{our identified supportive pretraining data is overall effective for ICL}, 
though the cross-task results show a portion of the ICL behavior can be task-specific and not universal across tasks. 


\begin{table*}[t]
    \centering
    \begin{tabular}{@{}l|p{0.58in}|p{0.58in}|p{0.58in}|p{0.58in}|p{0.58in}|p{0.58in}@{}}
    \toprule
     
     \multirow{2}{*}{{\diagbox{\emph{Source}}{\emph{Eval}}}} &SST-2& AG News & Story Cloze & SMS Spam & Sentiment 140 & TweetQA \\
     
      & 75.47  & 74.12 & 66.09 & 45.07 & 67.23 & 62.36\\
     
      \midrule
      SST-2 & \textcolor{violet}{\textbf{83.15} \qquad 75.87\tiny{$\pm$ 1.64}} & \textcolor{black}{\textbf{74.91} \qquad 73.24\tiny{$\pm$ 1.24}} & \textcolor{black}{\textbf{67.76} \qquad 66.24\tiny{$\pm$ 1.25}} & \textcolor{black}{52.48 \qquad 49.82\tiny{$\pm$ 4.50}} & \textcolor{black}{\textbf{69.03} \qquad 66.23\tiny{$\pm$ 1.24}} & \textcolor{black}{\textbf{62.20} \qquad 61.75\tiny{$\pm$ 0.26}}  \\
      \midrule
      AG News & \textcolor{black}{\textbf{79.04} \qquad 74.99\tiny{$\pm$ 0.77}} & \textcolor{violet}{\textbf{75.40} \qquad 73.77\tiny{$\pm$ 0.41}} & \textcolor{black}{\textbf{68.34} \qquad 66.38\tiny{$\pm$ 0.69}} & \textcolor{black}{\textbf{59.24} \qquad 46.55\tiny{$\pm$ 4.24}} & \textcolor{black}{\textbf{68.96} \qquad 66.23\tiny{$\pm$ 1.24}} & \textcolor{black}{61.86 \qquad 62.02\tiny{$\pm$ 0.55}}  \\
      \midrule
      Story Cloze & \textcolor{black}{\textbf{75.33} \qquad 72.50\tiny{$\pm$ 2.53}} & \textcolor{black}{74.12 \qquad 73.77\tiny{$\pm$ 0.41}} & \textcolor{violet}{\textbf{67.47} \qquad 65.25\tiny{$\pm$ 1.52}} & \textcolor{black}{51.36 \qquad 47.15\tiny{$\pm$ 4.90}} & \textcolor{black}{\textbf{69.92} \qquad 66.23\tiny{$\pm$ 1.24}} & \textcolor{black}{62.33 \qquad 62.02\tiny{$\pm$ 0.55}}  \\
      \midrule
      SMS Spam & \textcolor{black}{73.88 \qquad 75.87\tiny{$\pm$ 1.64}} & \textcolor{black}{72.78 \qquad 73.77\tiny{$\pm$ 0.41}} & \textcolor{black}{\textbf{67.25} \qquad 65.25\tiny{$\pm$ 1.52}} & \textcolor{violet}{\textbf{64.69} \qquad 46.55\tiny{$\pm$ 4.24}} & \textcolor{black}{63.70 \qquad 66.33\tiny{$\pm$ 1.34}} & \textcolor{black}{\textbf{62.13} \qquad 61.75\tiny{$\pm$ 0.26}}  \\
      \midrule
      Sentiment 140 & \textcolor{black}{\textbf{77.56} \qquad 73.49\tiny{$\pm$ 2.33}} & \textcolor{black}{72.78 \qquad 73.77\tiny{$\pm$ 0.41}} & \textcolor{black}{66.78 \qquad 66.38\tiny{$\pm$ 0.69}} & \textcolor{black}{\textbf{51.64} \qquad 44.52\tiny{$\pm$ 2.45}} & \textcolor{violet}{66.66 \qquad 66.00\tiny{$\pm$ 1.41}} & \textcolor{black}{\textbf{62.93} \qquad 61.64\tiny{$\pm$ 0.21}}  \\
      \midrule
      TweetQA & \textcolor{black}{\textbf{75.22} \qquad 72.50\tiny{$\pm$ 2.53}} & \textcolor{black}{71.52 \qquad 73.01\tiny{$\pm$ 1.42}} & \textcolor{black}{66.27 \qquad 64.91\tiny{$\pm$ 2.01}} & \textcolor{black}{43.09 \qquad 44.52\tiny{$\pm$ 2.45}} & \textcolor{black}{66.76 \qquad 66.33\tiny{$\pm$ 1.34}} & \textcolor{violet}{61.31 \qquad 61.33\tiny{$\pm$ 0.80}}  \\
      \bottomrule
    \end{tabular}
    \caption[Caption for LOF]{Evaluation of supportive pretraining data to ICL. We obtain supportive pretraining data using the guidance of a \emph{source} task and \emph{evaluate} ICL on all tasks. In the headers, we show the ICL performance of the original LM. We perform perturbative continued pretraining with both our selected supportive data (top number in cells) and an equal number of randomly sampled pretraining data (bottom number in cells). Diagonal cells indicate same-task evaluation and are marked purple. Our performance is bolded when the difference exceeds one standard deviation. On 4 of 6 tasks, the same-task ICL performance gain is observed (diagonal). On 5 of 6 tasks, the corresponding supportive pretraining data improves ICL on at least three tasks (rows). 
    }
    \label{tab:main_result}
\end{table*}

\paragraph{Control evaluation on zero-shot data} 
Being effective on the ICL data does not necessarily mean a direct support for a model's ICL ability, which is to learn from the demonstration examples. The test input can be a confounding factor: if our selection is effective as well on zero-shot test input without demonstrations, then the selection is not specific to the ICL ability. 
Therefore, we further confirm the supportiveness of our selected supportive pretraining data to ICL, contrastively in a zero-shot setup. 
We evaluate our models after perturbative continued pretraining in \autoref{tab:main_result} on the same tasks but without the in-context demonstrations. 
We present the results in \autoref{tab:nc_results}. The two columns show the zero-shot prompting performance of the original LM and the model after continued pretraining with our ICL-supportive selection, respectively. 
We do not observe performance gain for most tasks, indicating \textbf{our selection is specific to the ICL ability without benefiting the zero-shot, no-demonstration task performance}.

\begin{table}[ht]
    \centering
    \begin{tabular}{@{}l|p{0.6in}p{0.85in}@{}}
    \toprule
      Zero-shot Eval & Original & +\small{ICL-supportive} \\
      \midrule
      SST-2 & 46.82 & 46.83 \\
      [2pt]
      AG News & 46.14 & 44.05 \\
      [2pt]
      Story Cloze & 50.43 & 51.39 \\
      [2pt]
      SMS Spam & 44.41 & 43.84 \\
      [2pt]
      Sentiment 140 & 55.84 & 54.90 \\
      [2pt]
      TweetQA & 50.44 & 50.32 \\
      \bottomrule
    \end{tabular}
    \caption[Caption for LOF]{Control evaluation. We report the zero-shot prompting performance of the original LM and the perturbed LM after trained on our selected supportive pretraining data. No significant performance gain is observed for most tasks, showing our selected supportive pretraining data is specific to ICL without improving the zero-shot, no-demonstration task performance. 
    }
    \label{tab:nc_results}
\end{table}


%% file: sections/analysis.tex

\label{sec:analysis}

In the previous section, we identify a small subset of pretraining data that supports the ICL ability of language models. In this section, we analyze the selected supportive pretraining data to understand what makes them useful to ICL. Specifically, we compare the supportive pretraining data contrastively with randomly sampled pretraining instances, investigating three aspects of the pretraining data: the domain relevance to downstream tasks, the token frequency distribution, and the information gain of incorporating long-range context. 




\subsection{Domain relevance}
\label{sec:domain_sec}

\citet{xie2022an} and \citet{min2022rethinking} imply that in-context demonstration is useful since it helps locate a particular domain or concept of the test input the LM already learned through the pretraining data. On the other hand, \citet{olsson2022context} imply that in-context demonstration is useful because the decision over the test input may be done through a soft-copy mechanism from the demonstration examples. These lead to two different expectations of the role of supportive pretraining data: 
(1) Inferred from \citet{xie2022an} and \citet{min2022rethinking}, the supportive pretraining data should be from a same domain as the demonstration and test examples, providing direct supporting knowledge to solve the downstream task. 
(2) Inferred from \citet{olsson2022context}, the supportive pretraining data should be beneficial to the soft-copy mechanism, providing meta support for the abstract ability, unconstrained with the concrete data domain.\footnote{This view of supportive data will be revisited in \Sref{sec:ig}.} 
We aim to measure the domain relevance between supportive pretraining data and downstream tasks. 

\paragraph{Method} 
To quantify domain relevance, we use MAUVE score \citep{Pillutla2021MAUVEMT} to measure an information divergence between two text distributions. We compute two MAUVE scores, between the target task data 
and our selected supportive pretraining data, and between the task data and random pretraining data. We then compute and report their difference. 
A positive MAUVE difference indicates a higher domain relevance of our supportive pretraining data.\footnote{\citet{Pillutla2021MAUVEMT} also shows higher MAUVE indicates higher generation quality, but we skip that aspect since all of our data are naturally occuring text.} We use RoBERTa \citep{liu2019roberta} as MAUVE's embedding model following \citet{he2022blind}. 

\paragraph{Results}

\begin{figure}[t]
    \centering
    \includegraphics[width=0.46\textwidth]{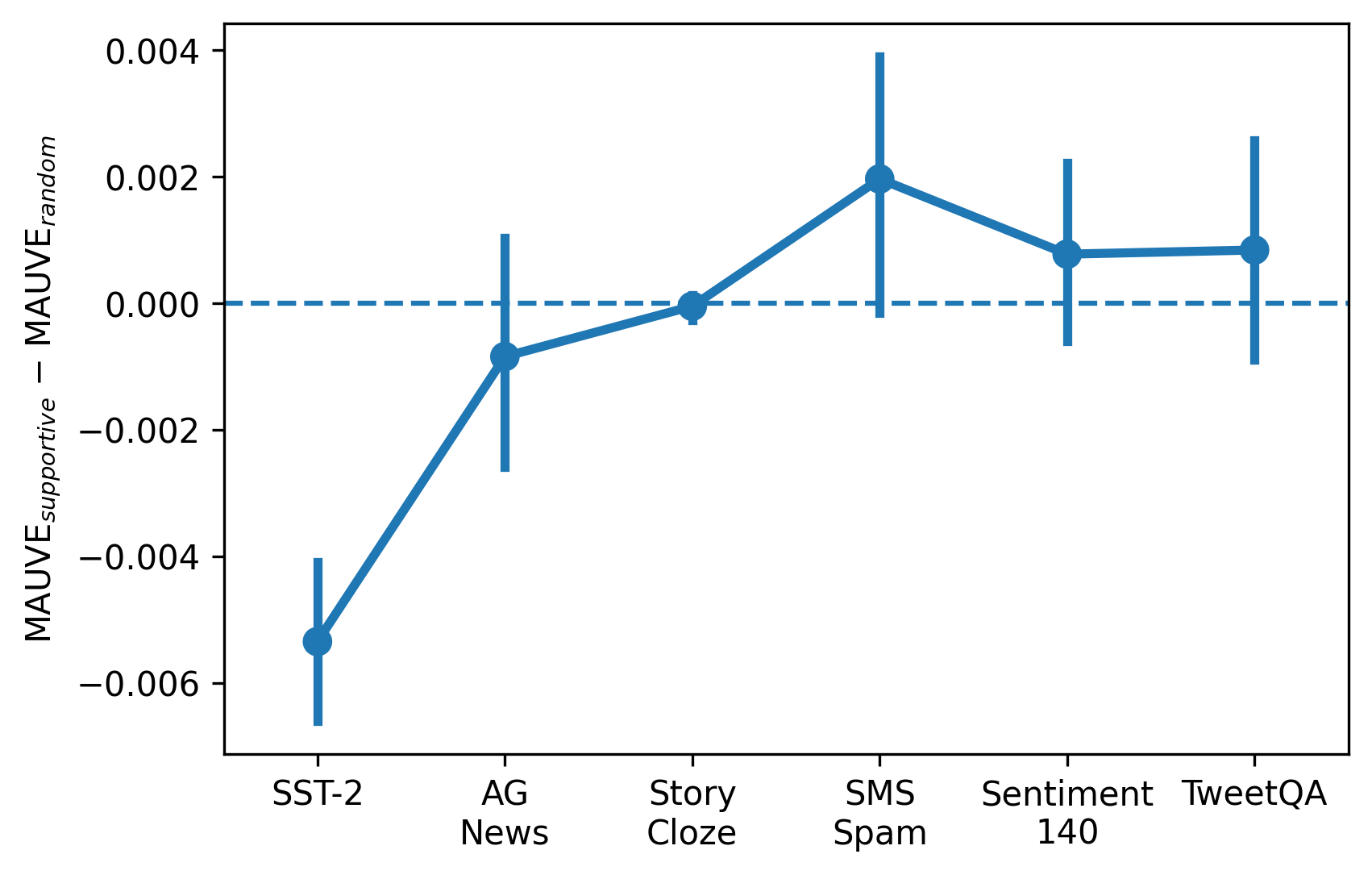}
    \caption{The MAUVE score between the supportive pretraining data and target task data, subtracted by the MAUVE score between random data and target task data. The error bars indicate the 95\% confidence interval. No tasks show the supportive data has a significant higher domain relevance compared to random data. 
    }
    \label{fig:mauve}
\end{figure}

We show the difference of MAUVE scores in \autoref{fig:mauve}. 
The error bar shows the 95\% confidence interval using 32 random seeds. We find that for 5 of the 6 tasks, there is no significant difference between the MAUVE scores of supportive pretraining data and random data. For SST-2, the supportive pretraining data even shows a lower MAUVE score. Therefore, \textbf{the supportive pretraining data to ICL do \emph{not} have a higher domain relevance to the task, compared to general pretraining data}. 
This result aligns with the domain relevance finding in \citet{shin2022effect} where dataset-level analyses were performed. 
This implies the improved ICL behavior of our models may be a meta ability, aided by pretraining data unrelated to the specific domain knowledge for solving the task, but related to a domain-invariant mechanism to learn from a data's context. \Sref{sec:ig} continues this discussion.

\subsection{Token frequency distribution} 
Providing demonstrations to a task input under an ICL setup creates repetitions (e.g., of label tokens), which changes the token frequency distribution of the ICL task data. Therefore, we are interested in whether the supportive pretraining data possess a different token frequency distribution from general pretraining data. 
Experimented with sequences of image-label pairs, \citet{chan_data_2022} find that a skewed class distribution (high burstiness) 
and a large number of rarely occurring classes in training data promote the ICL ability of Transformer models \citep{vaswani2017attention}. 
However, it is unknown whether the findings on the synthetic image-label data can transfer to the natural language pretraining data, a gap we address in this subsection. 


\paragraph{Method}
We fit a Zipfian distribution over each supportive and random pretraining instance that consists of 2048 tokens. The Zipf's coefficient is the negative slope of a linear regression over the tokens' log-rank v.s. log-frequency. A higher Zipf's coeffcient indicates a higher mass on the frequent tokens (i.e., more skewed distribution). A lower Zipf's coefficient indicates a higher mass on the rare, long-tail tokens (i.e., flatter distribution). 


\paragraph{Results}
\begin{figure}[t]
    \centering
    \includegraphics[width=0.46\textwidth]{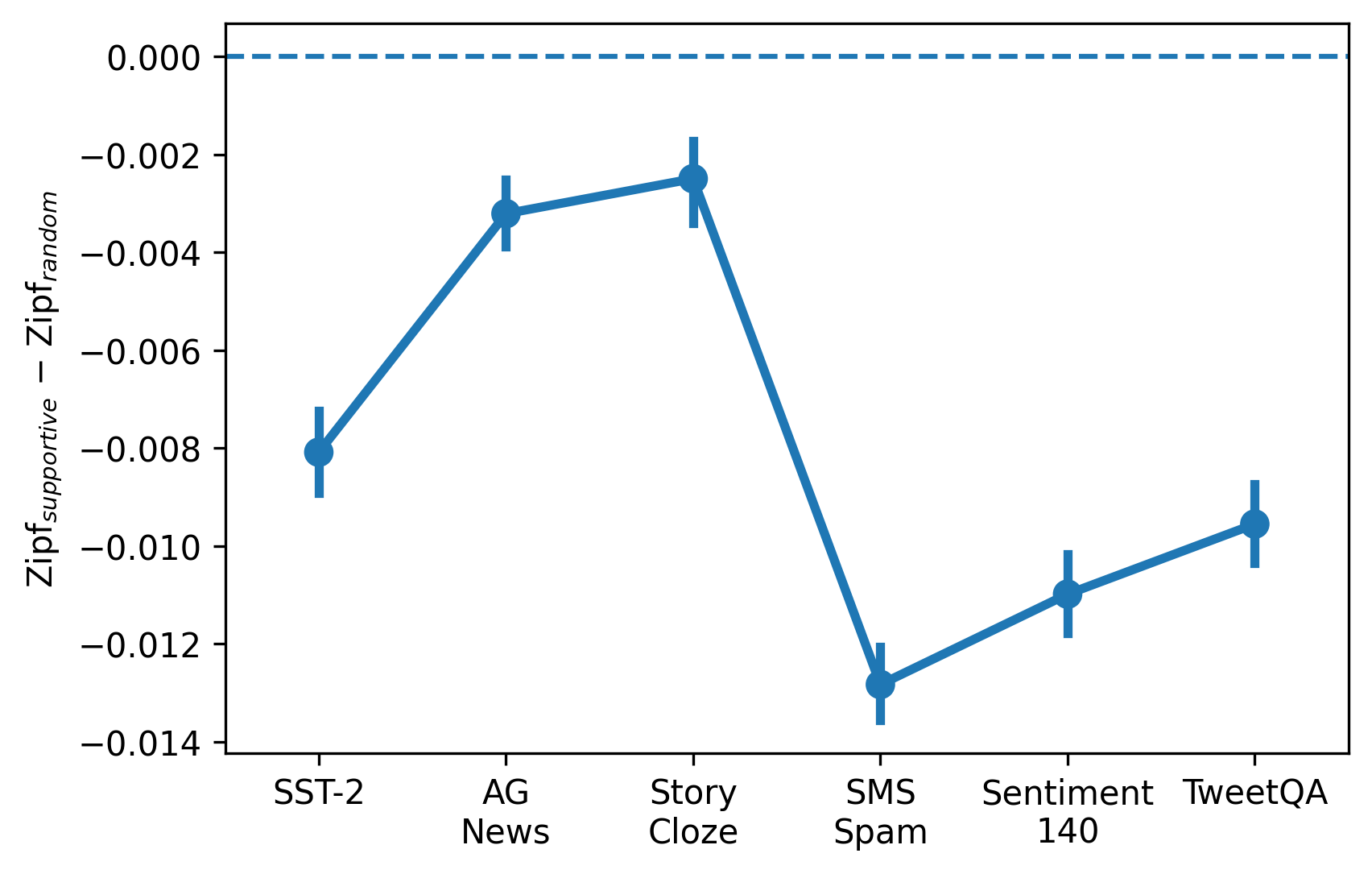}
    \caption{The difference in average Zipf's coefficients of the token frequency distribution of supportive pretraining instances and random examples. 
    The error bars indicate the 95\% confidence interval. We find a lower Zipf's coefficient for supportive pretraining data, indicating a flatter frequency distribution, with a relatively higher mass on the rare, long-tail tokens. 
    }
    \label{fig:zipf}
\end{figure}

In \autoref{fig:zipf}, we show the difference in average Zipf's coefficients between supportive and random pretraining data, each with a group size of 2000. The error bar shows the 95\% confidence interval with 32 random seeds. We find that for all tasks, the Zipf's coefficient of the supportive pretraining data is significantly \emph{lower} than that of the random pretraining data. This indicates a flatter Zipfian distribution with a relatively higher mass over the long-tail tokens. In other words, though the overall burstiness of data is lower, \textbf{there is a relatively higher amount of rarely occurring, long-tail tokens in the supportive pretraining data for ICL}. 
Flatter frequency distribution also indicates higher entropy over the tokens, presumably making the supportive pretraining data \emph{challenging} examples to fit by the model, a concept we explore further in the next subsection.

\begin{figure*}[t]
    \centering
    \begin{subfigure}[h]{0.3\textwidth}
        \includegraphics[width=\textwidth]{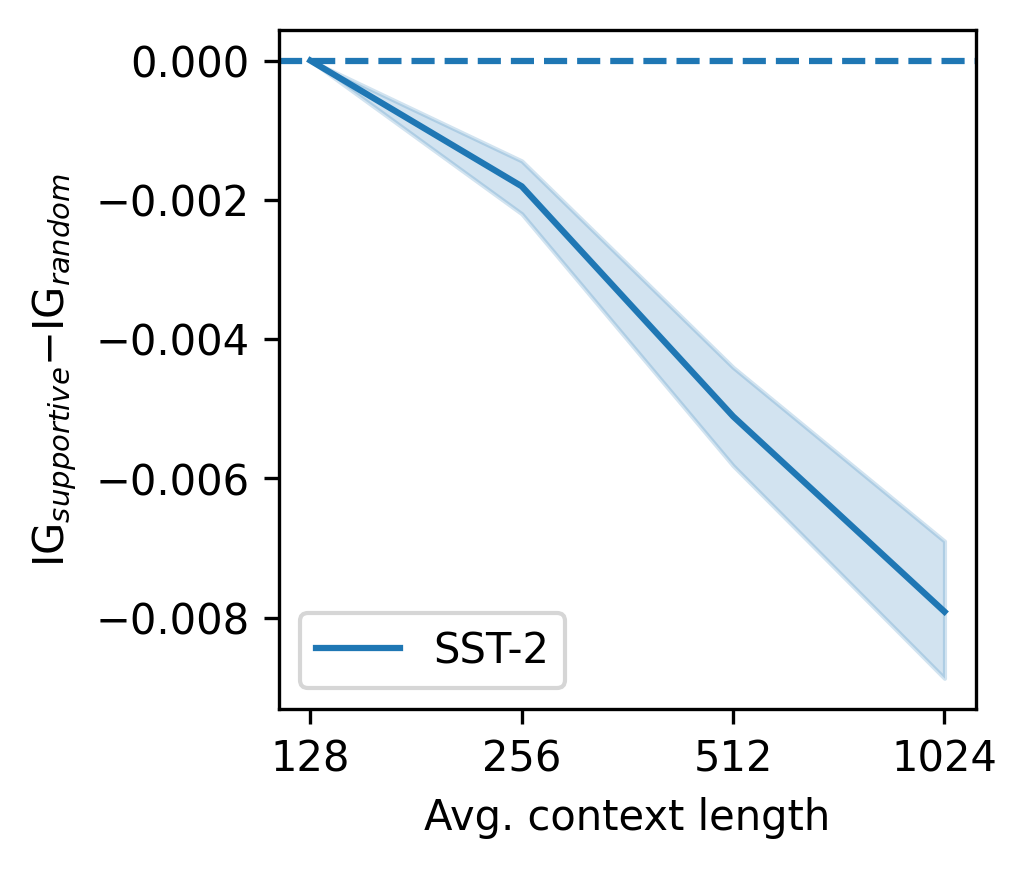}
        \label{fig:ics1}
    \end{subfigure}
    \hfill
    \begin{subfigure}[h]{0.3\textwidth}
        \includegraphics[width=\textwidth]{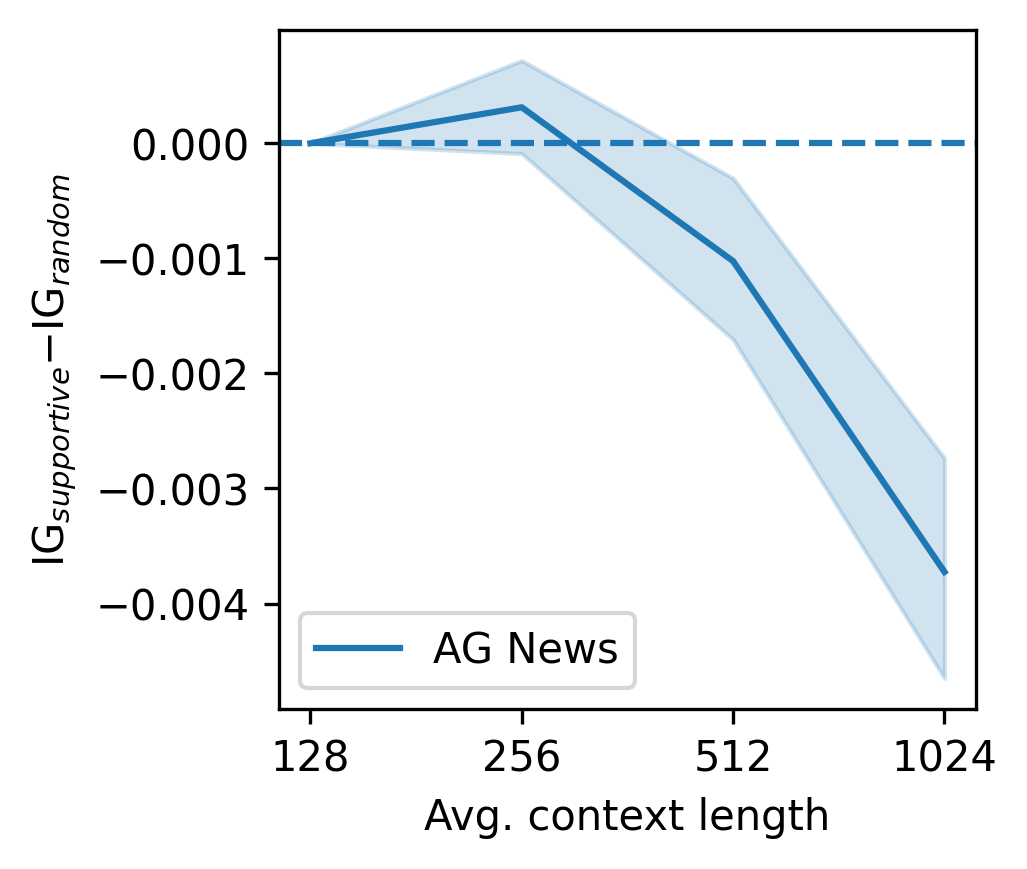}
        \label{fig:ics2}
    \end{subfigure}
    \hfill
    \begin{subfigure}[h]{0.3\textwidth}
        \includegraphics[width=\textwidth]{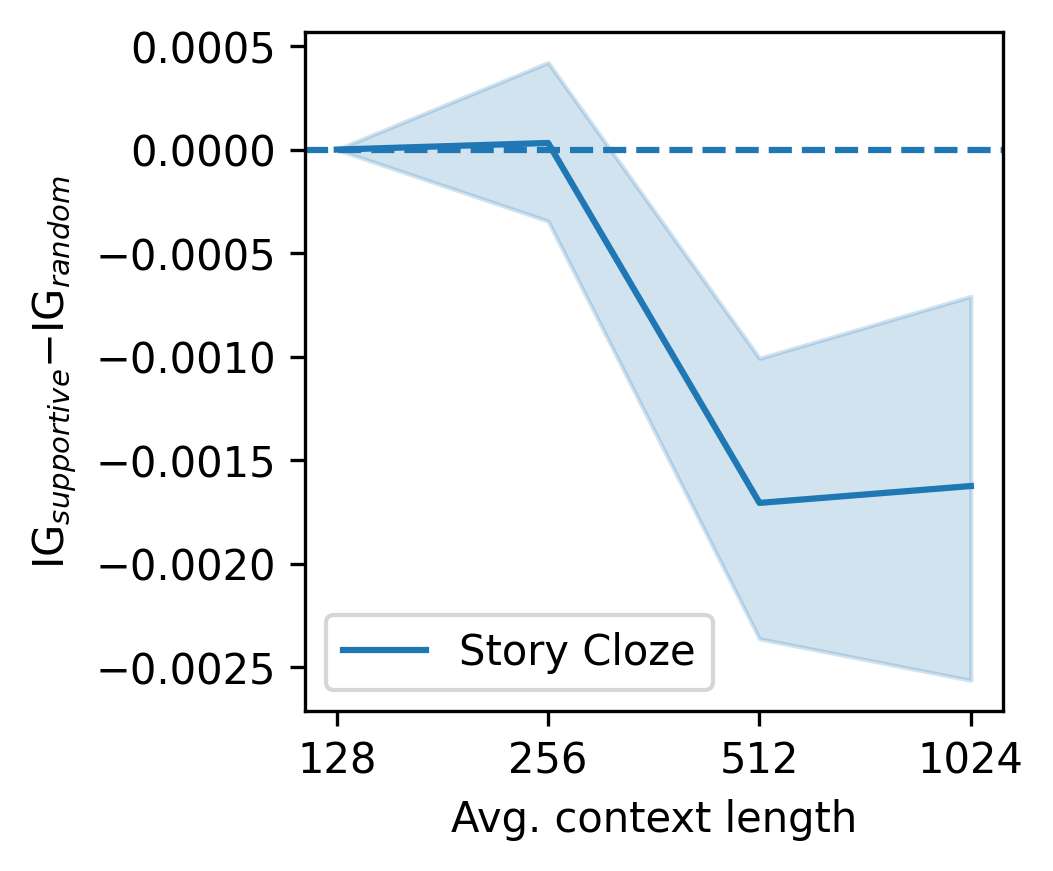}
        \label{fig:ics3}
    \end{subfigure}
    \vfill
    
    \vspace{-2.7em}
    
    \begin{subfigure}[h]{0.3\textwidth}
        \includegraphics[width=\textwidth]{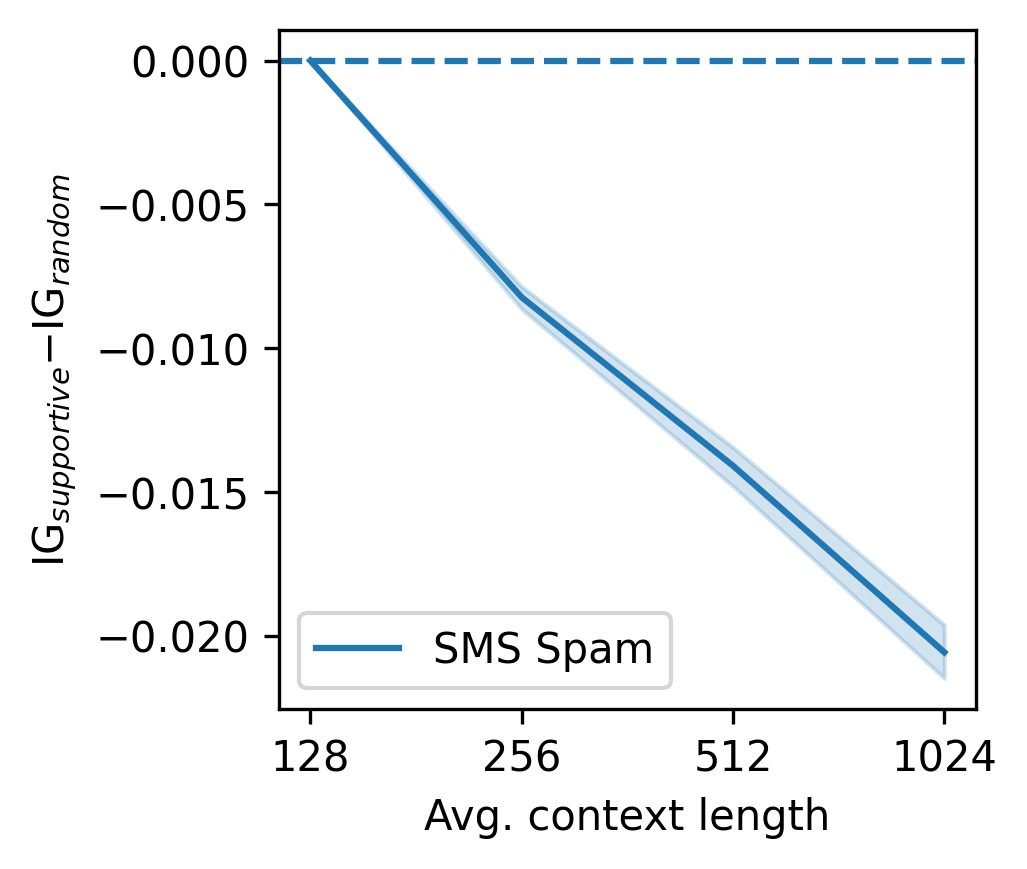}
        \label{fig:ics4}
    \end{subfigure}
    \hfill
    \begin{subfigure}[h]{0.3\textwidth}
        \includegraphics[width=\textwidth]{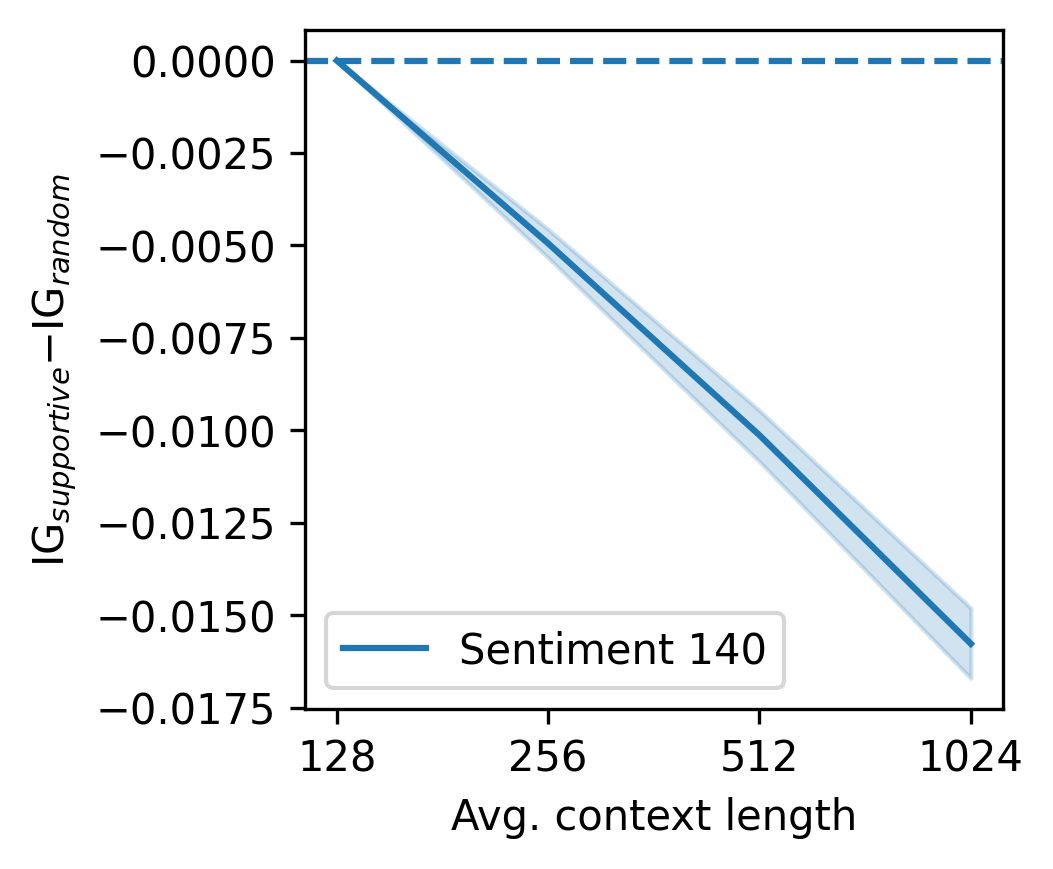}
        \label{fig:ics5}
    \end{subfigure}
    \hfill
    \begin{subfigure}[h]{0.3\textwidth}
        \includegraphics[width=\textwidth]{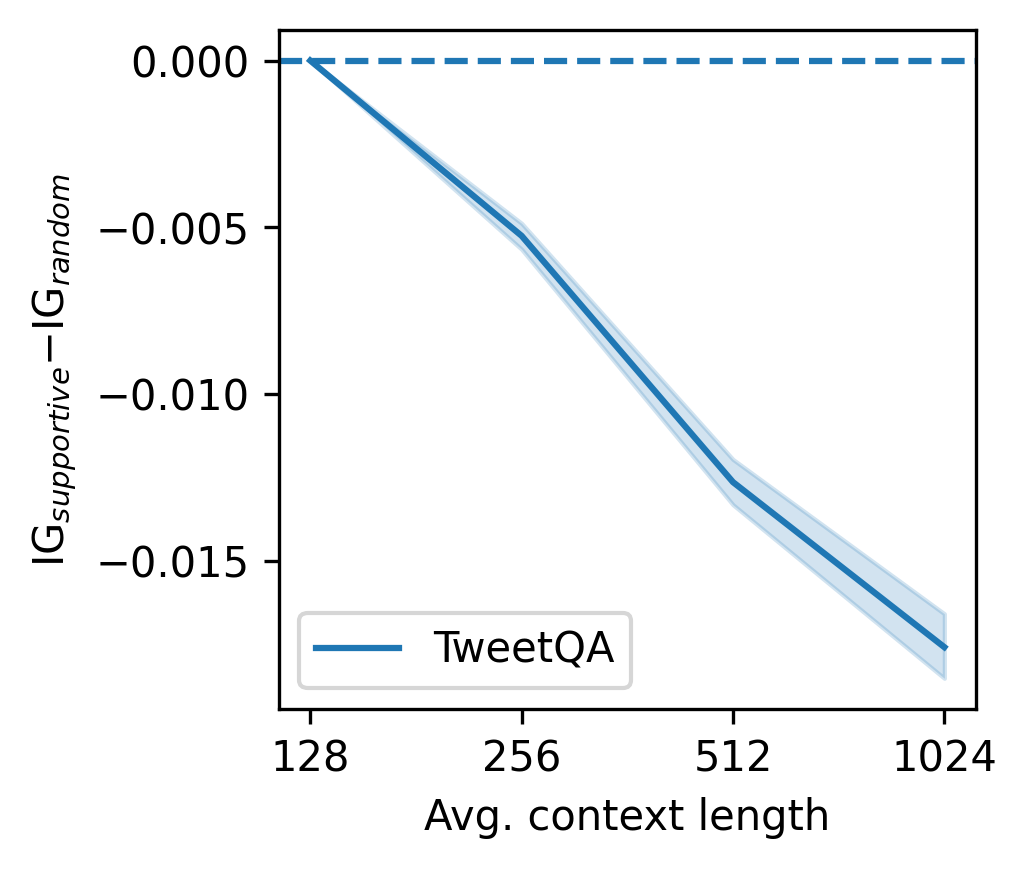}
        \label{fig:ics6}
    \end{subfigure}
    
    
    \vspace{-1.5em}
    \caption{
    The difference between supportive pretraining instances and random examples in information gain of incorporating long-range context for next-token prediction. 
    We fix the average short context length ($s$) at 128 tokens and iterate through long context lengths ($l$) of \{256, 512, 1024\}. The shaded area shows the 95\% confidence interval. 
    The results show that the long-range context in supportive pretraining data leads to a lower information gain than random pretraining examples. Supportive pretraining data are \emph{challenging} examples in incorporating their long-range context.  
    }
    \label{fig:ig}
\end{figure*}

\subsection{Information gain from long-range context} 
\label{sec:ig}

In \Sref{sec:domain_sec}, we find that the domain relevance of the supportive pretraining data to downstream tasks is not higher than that of random pretraining data. 
This is comprehendible if we follow the aforementioned perspective of \citet{olsson2022context}, hypothesizing that there exists a soft-copy mechanism between the in-context demonstrations and test input. The supportive pretraining data may provide meta support for the abstract soft-copy mechanism rather than task-specific knowledge. 
We further hypothesize that to facilitate such meta support, the incorporation of long-range context during language modeling in supportive pretraining data should be different from random pretraining data, since the demonstration examples in the ICL setup is a form of long-range context. 
We propose a novel information gain measure to quantify this feature of incorporating long-range context.

\paragraph{Method}
Recall that the canonical definition of information gain (IG) is $\mathrm{IG}(T, a) = H(T) - H(T \mid a)$, where $T$ is a target variable, $a$ is an attribute conditioned on by $T$, and $H(\cdot)$ computes entropy. It measures the decrease of entropy (thus the gain of information) in $T$ if conditioned on $a$. We adapt the canonical IG to measure the decrease of cross entropy for each token ($w_i$) in a pretraining dataset when conditioned on a long ($l$) context over a short ($s$) context: 
\begin{align*}
    \mathrm{IG}(l, s) = \mathrm{CE}(w_i \mid \text{ctx}_s) - \mathrm{CE}(w_i \mid \text{ctx}_l)
\end{align*}
Ideally the length of long or short context should remain constant across different tokens $w_i$, but it would be a very expensive computation due to a lack of parallelism. We approximate the computation by splitting a full sequence of pretraining tokens (e.g., 2048 tokens) to smaller blocks and calculate cross entropy with the boundary of blocks: 
\begin{align*}
    \mathrm{IG}(l, s) =& - \log p_{\theta}(w_i \mid w_{i-(i \;\mathrm{mod}\; 2s)~:~i})\\
    & + \log p_{\theta}(w_i \mid w_{i-(i \;\mathrm{mod}\; 2l)~:~i})
\end{align*}
With the above definition, the average length of context for all $w_i$ is $s$ and $l$, respectively. In the experiments below, we keep $s=128$ for the length of short context and increase the length of long context at $l=\{256, 512, 1024\}$. 

We report the difference in the average information gain (across $w_i$) of incorporating long-range context for a language modeling objective, in supportive pretraining data over random pretraining data. Additionally, we want to use the defined information gain measure as a standalone feature of data, so we use a different LM to compute the cross entropy than the LM on which we perform ICL. Below we report results using OPT-1.3B, while experiments using OPT-350M shows a similar trend. 

\paragraph{Results}
In \autoref{fig:ig}, we see for all of the experimented tasks, there is a significant trend that increasing the length $l$ for the long-range context for supportive pretraining data has a \emph{lower} relative information gain compared to random pretraining data. 
Though seeming counterintuitive at first glance, this suggests that \textbf{the supportive pretraining data are more \emph{challenging} examples in incorporating the long-range context information}.\footnote{Note that a reverse of the statement may not hold necessarily, since an example's long-range context can also be irrelevant by nature and challenging in a useless way.} 
A possible explanation for this is that such challenging examples contain confounding spans that harms the information gain measure. The language model has to learn to decide which part of the long-range context is truly relevant to the prediction of next tokens. This would resemble more and thus helpful to the ICL task scenario where there are multiple demonstrations from different classes. 


\subsection{Future work}
\label{sec:discussion}


Despite our aforementioned findings, we mainly conduct correlational analyses throughout the work. Despite the potential confounding factors, future work can try converting the correlational findings to causal ones. For example, to actively refine or construct pretraining data to improve existing models' ICL performance, with a metric of token frequency distribution (i.e., find data with a higher mass of long-tail tokens) or context information gain (i.e., find difficult examples in incorporating long-range context). Additionally, we only investigate classification tasks in this work. However, the ORCA-ICL method can be applicable to generation tasks as well in the future, if the ICL loss is defined over a sequence probability of the generation.



%% file: sections/related_work.tex

\paragraph{Demonstration examples}~\citet{min2022rethinking} understand ICL through analyzing which aspects of the demonstration examples contribute or are irrelevant to task performance. They find replacing ground truth demonstration labels with random labels would not hurt task performance, while ICL still benefits from knowing the label space, distribution of inputs, and sequence format specified in demonstration examples.\footnote{Recent work like \citet{Wei2023LargerLM} and \citet{Pan2023WhatIL} show the related findings would depend on model scales as well.} 
\citet{zhang2022robustness} further show on sequence labeling tasks, the length of demonstrations and the relevance of their tokens are important for ICL. 

\paragraph{Learning mechanism} 
\citet{xie2022an} explain ICL as implicit Bayesian inference, occurring when language models infer a shared latent concept from demonstration examples at inference time. They show language models exhibit such ICL behavior by constructing synthetic pretraining data with a controlled distribution of concepts. 
\citet{garg2022can} empirically show that Transformer models can be trained to learn unseen linear functions from in-context demonstration examples. 
\citet{olsson2022context} present evidence that multi-layer attention-based models form an induction head and perform ICL by a pattern copying behavior from the prefixing context. 
More recent work like \citet{akyurek2022learning}, \citet{dai2022can}, and \citet{von2022transformers} explain ICL in Transformer models as a kind of standard learning algorithms over the demonstration examples, such as gradient descent and regression. 

\paragraph{Pretraining data} 

\citet{Razeghi2022ImpactOP} find on numerical reasoning tasks, a language model's ICL performance is highly correlated with the term frequency of the input data in the pretraining corpus. 
\citet{shin2022effect} investigate how ICL can be affected when the pretraining dataset varies. They discover that ICL heavily depends on the corpus domain source, but pretraining with a corpus related to a downstream task does not always translate to a competitive ICL performance on the task.  
\citet{chan_data_2022} experiment on a synthetic image-label pairs dataset. They show certain distributional properties of the synthetic pretraining data, such as the burstiness of classes and large numbers of rarely occurring classes, promote the emergence of ICL. 
Our work belongs to this line of work, but offers a first step towards understanding ICL in realistic NLP tasks through analyzing instance-level pretraining data. 
Additionally, concurrent to our work, \citet{Gu2023PreTrainingTL} propose a method that groups pretraining data by their instrinsic tasks, enhancing instead of interpreting existing language models' ICL ability.


%% file: sections/conclusion.tex
In-context learning has shown superior performance on a range of NLP tasks, yet it remained unclear \emph{from where} language models acquired this ability. We approach the problem by identifying a small subset of pretraining data that particularly supports language models to do in-context learning on downstream tasks. 
We analyze common features of the supportive instances in contrast to general pretraining data and find that: (1) The supportive pretraining data do \emph{not} have a higher domain relevance to the downstream tasks. (2) The supportive data contain a relatively larger amount of rare, long-tail tokens. (3) The supportive pretraining data are more \emph{challenging} instances in incorporating long-range context in language modeling. Our findings may be beneficial to future work that refine or construct pretraining data, in order to actively improve existing models' in-context learning performance.

%% file: sections/limitation.tex


It is worth noting that the supportive pretraining data we investigated throughout the work is w.r.t. the \emph{current} LM, such that a perturbative continued pretraining with the supportive data would improve the final LM checkpoint deployed to downstream tasks. It is possible that for some data which we did not determine as supportive, they \emph{had been} supportive w.r.t. early checkpoints of the LM. With more computing resources, future work may investigate the trend of supportive patterns across multiple checkpoints of a LM throughout the pretraining process.

Additionally, another significant limitation of our work is the amount of involved computing resource. 
The ORCA-ICL method is gradient-based that requires back-propagation. Since we iterate through a large size of pretraining data, the cost of computation is similar to training a language model with a batch size of 1 on the considered pretraining data. 
On our 4 nodes each consists of 8 Nvidia V100 GPUs, finding the supportive pretraining data for \emph{each} source task in our experiment would take about a week. 
One mitigating aspect of such computation is that the gradient calculation can be done asynchronously, therefore enabling the use of idle, leftover GPUs scattered across a cluster of nodes. 
We plan to explore efficient computation of gradient similarity or move from a paradigm of extracting supportive data to generating supportive data in future work.

%% file: sections/appendix.tex


\section{Qualitative examples}
In \autoref{tab:qual_ex}, we show some qualitative examples of the supportive pretraining data to ICL and random pretraining data. Note that these are illustrative examples extracted from long pretraining instances (each instance consists of 2048 tokens), for a better understandability of our findings. A manual examination of such data is difficult, and we thus propose the quantitative analyses described in the main paper. 

\begin{table}[t]
\begin{center}
\renewcommand{\arraystretch}{1.0}
\begin{tabular}{p{0.46\textwidth}}
    \toprule
    Supportive pretraining data to ICL\\
    {\begin{lstlisting}
...
Samsung's new Odyssey+ headset could fix its muddled VR vision
As one of the world's most technologically innovative companies, Samsung should be leading the pack in VR - one of the decade's top transformative technologies. Instead, it has largely let Microsoft and Facebook determine its role in the VR space, leading to its current situation as an also-ran.
If I was betting on whether that will change anytime soon, an FCC leak of the company's new Odyssey+ VR headset (discovered by RoadtoVR) would point to "no." Most of the specs are staying the same as its prior, Windows-dependent Odyssey model: Each eye still gets a 3.5-inch screen with 1,440 by 1,600 resolution, combining for a 110-degree field of view, and AMOLED technology will be used to guarantee dark blacks and rich colors.
There's one mystery in the new specs, namely a reference to the AMOLED screens now including something called "SFS."
...
    \end{lstlisting}}\\
    \midrule
    Random pretraining data\\
    {\begin{lstlisting}
...
Bangladesh authorities and intelligence officials have long been saying that many of the refugees are involved in illicit drug trade, smuggling, robbery and ransom-seeking. Earlier Tuesday, the elite security agency Rapid Action Battalion arrested nine refugees suspected of being involved in various criminal activities.
They had firearms, bullets and sharp weapons, Islam said. Local media reported that Tuesday's chaos began after the arrest of the suspects as one group blamed another for helping the security agency in detaining them. Human rights groups that are involved in the camps acknowledge there are criminal elements among the Rohingya refugees.
...
    \end{lstlisting}}\\
    \bottomrule
\end{tabular} 
\end{center}
\caption{
Qualitative examples of the supportive pretraining data to ICL in the task of SMS spam detection. We also show an example of random pretraining data for comparison. As our finding on domain relevance suggested, neither of the examples are about SMS spam, so the language model may not learn direct knowledge about the task from supportive pretraining data to ICL. Compared to the random data, the supportive data to ICL has some relatively low-frequency tokens appear multiple times (e.g., VR, Odyssey, AMOLED) and the language model may learn some meta-knowledge about ICL (e.g., copying behaviors from the context) based on them. However, such patterns are sparse, noisy, and hard to analyze through manual inspections. We therefore present the quantitative analyses in the main paper. 
}
\label{tab:qual_ex}
\end{table}

%% file: paper.bbl
\begin{thebibliography}{41}
\expandafter\ifx\csname natexlab\endcsname\relax\def\natexlab#1{#1}\fi

\bibitem[{Aky{\"u}rek et~al.(2022)Aky{\"u}rek, Schuurmans, Andreas, Ma, and
  Zhou}]{akyurek2022learning}
Ekin Aky{\"u}rek, Dale Schuurmans, Jacob Andreas, Tengyu Ma, and Denny Zhou.
  2022.
\newblock What learning algorithm is in-context learning? investigations with
  linear models.
\newblock \emph{arXiv preprint arXiv:2211.15661}.

\bibitem[{Almeida et~al.(2011)Almeida, Hidalgo, and
  Yamakami}]{almeida2011contributions}
Tiago~A Almeida, Jos{\'e} Mar{\'\i}a~G Hidalgo, and Akebo Yamakami. 2011.
\newblock Contributions to the study of sms spam filtering: new collection and
  results.
\newblock In \emph{Proceedings of the 11th ACM symposium on Document
  engineering}, pages 259--262.

\bibitem[{Brown et~al.(2020)Brown, Mann, Ryder, Subbiah, Kaplan, Dhariwal,
  Neelakantan, Shyam, Sastry, Askell, Agarwal, Herbert-Voss, Krueger, Henighan,
  Child, Ramesh, Ziegler, Wu, Winter, Hesse, Chen, Sigler, Litwin, Gray, Chess,
  Clark, Berner, McCandlish, Radford, Sutskever, and
  Amodei}]{NEURIPS2020_1457c0d6}
Tom Brown, Benjamin Mann, Nick Ryder, Melanie Subbiah, Jared~D Kaplan, Prafulla
  Dhariwal, Arvind Neelakantan, Pranav Shyam, Girish Sastry, Amanda Askell,
  Sandhini Agarwal, Ariel Herbert-Voss, Gretchen Krueger, Tom Henighan, Rewon
  Child, Aditya Ramesh, Daniel Ziegler, Jeffrey Wu, Clemens Winter, Chris
  Hesse, Mark Chen, Eric Sigler, Mateusz Litwin, Scott Gray, Benjamin Chess,
  Jack Clark, Christopher Berner, Sam McCandlish, Alec Radford, Ilya Sutskever,
  and Dario Amodei. 2020.
\newblock \href
  {https://proceedings.neurips.cc/paper/2020/file/1457c0d6bfcb4967418bfb8ac142f64a-Paper.pdf}
  {Language models are few-shot learners}.
\newblock In \emph{Advances in Neural Information Processing Systems},
  volume~33, pages 1877--1901. Curran Associates, Inc.

\bibitem[{Chan et~al.(2022)Chan, Santoro, Lampinen, Wang, Singh, Richemond,
  McClelland, and Hill}]{chan_data_2022}
Stephanie~CY Chan, Adam Santoro, Andrew~Kyle Lampinen, Jane~X Wang, Aaditya~K
  Singh, Pierre~Harvey Richemond, James McClelland, and Felix Hill. 2022.
\newblock Data distributional properties drive emergent in-context learning in
  transformers.
\newblock In \emph{Advances in Neural Information Processing Systems}.

\bibitem[{Chowdhery et~al.(2022)Chowdhery, Narang, Devlin, Bosma, Mishra,
  Roberts, Barham, Chung, Sutton, Gehrmann et~al.}]{chowdhery2022palm}
Aakanksha Chowdhery, Sharan Narang, Jacob Devlin, Maarten Bosma, Gaurav Mishra,
  Adam Roberts, Paul Barham, Hyung~Won Chung, Charles Sutton, Sebastian
  Gehrmann, et~al. 2022.
\newblock Palm: Scaling language modeling with pathways.
\newblock \emph{arXiv preprint arXiv:2204.02311}.

\bibitem[{Dai et~al.(2022)Dai, Sun, Dong, Hao, Sui, and Wei}]{dai2022can}
Damai Dai, Yutao Sun, Li~Dong, Yaru Hao, Zhifang Sui, and Furu Wei. 2022.
\newblock Why can gpt learn in-context? language models secretly perform
  gradient descent as meta optimizers.
\newblock \emph{arXiv preprint arXiv:2212.10559}.

\bibitem[{Devlin et~al.(2019)Devlin, Chang, Lee, and
  Toutanova}]{Devlin2019BERTPO}
Jacob Devlin, Ming-Wei Chang, Kenton Lee, and Kristina Toutanova. 2019.
\newblock {BERT}: Pre-training of deep bidirectional transformers for language
  understanding.
\newblock In \emph{Proc. NAACL-HLT}.

\bibitem[{Dong et~al.(2022)Dong, Li, Dai, Zheng, Wu, Chang, Sun, Xu, and
  Sui}]{dong2022survey}
Qingxiu Dong, Lei Li, Damai Dai, Ce~Zheng, Zhiyong Wu, Baobao Chang, Xu~Sun,
  Jingjing Xu, and Zhifang Sui. 2022.
\newblock A survey for in-context learning.
\newblock \emph{arXiv preprint arXiv:2301.00234}.

\bibitem[{Garg et~al.(2022)Garg, Tsipras, Liang, and Valiant}]{garg2022can}
Shivam Garg, Dimitris Tsipras, Percy Liang, and Gregory Valiant. 2022.
\newblock What can transformers learn in-context? a case study of simple
  function classes.
\newblock \emph{arXiv preprint arXiv:2208.01066}.

\bibitem[{Go et~al.(2009)Go, Bhayani, and Huang}]{go2009twitter}
Alec Go, Richa Bhayani, and Lei Huang. 2009.
\newblock Twitter sentiment classification using distant supervision.
\newblock \emph{CS224N project report, Stanford}, 1(12):2009.

\bibitem[{Gu et~al.(2023)Gu, Dong, Wei, and Huang}]{Gu2023PreTrainingTL}
Yuxian Gu, Li~Dong, Furu Wei, and Minlie Huang. 2023.
\newblock Pre-training to learn in context.

\bibitem[{Guu et~al.(2023)Guu, Webson, Pavlick, Dixon, Tenney, and
  Bolukbasi}]{Guu2023SimfluenceMT}
Kelvin Guu, Albert Webson, Elizabeth-Jane Pavlick, Lucas Dixon, Ian Tenney, and
  Tolga Bolukbasi. 2023.
\newblock Simfluence: Modeling the influence of individual training examples by
  simulating training runs.
\newblock \emph{ArXiv}, abs/2303.08114.

\bibitem[{Han and Tsvetkov(2022)}]{han2022orca}
Xiaochuang Han and Yulia Tsvetkov. 2022.
\newblock Orca: Interpreting prompted language models via locating supporting
  data evidence in the ocean of pretraining data.
\newblock \emph{arXiv preprint arXiv:2205.12600}.

\bibitem[{He et~al.(2022)He, Zhang, Wang, Kumar, Cho, Glass, and
  Tsvetkov}]{he2022blind}
Tianxing He, Jingyu Zhang, Tianle Wang, Sachin Kumar, Kyunghyun Cho, James
  Glass, and Yulia Tsvetkov. 2022.
\newblock On the blind spots of model-based evaluation metrics for text
  generation.
\newblock \emph{arXiv preprint arXiv:2212.10020}.

\bibitem[{Hoffmann et~al.(2022)Hoffmann, Borgeaud, Mensch, Buchatskaya, Cai,
  Rutherford, Casas, Hendricks, Welbl, Clark et~al.}]{hoffmann2022training}
Jordan Hoffmann, Sebastian Borgeaud, Arthur Mensch, Elena Buchatskaya, Trevor
  Cai, Eliza Rutherford, Diego de~Las Casas, Lisa~Anne Hendricks, Johannes
  Welbl, Aidan Clark, et~al. 2022.
\newblock Training compute-optimal large language models.
\newblock \emph{arXiv preprint arXiv:2203.15556}.

\bibitem[{Holtzman et~al.(2019)Holtzman, Buys, Du, Forbes, and
  Choi}]{holtzman2019curious}
Ari Holtzman, Jan Buys, Li~Du, Maxwell Forbes, and Yejin Choi. 2019.
\newblock The curious case of neural text degeneration.
\newblock In \emph{International Conference on Learning Representations}.

\bibitem[{Kingma and Ba(2014)}]{kingma2014adam}
Diederik~P Kingma and Jimmy Ba. 2014.
\newblock Adam: A method for stochastic optimization.
\newblock \emph{arXiv preprint arXiv:1412.6980}.

\bibitem[{Koh and Liang(2017)}]{Koh2017UnderstandingBP}
Pang~Wei Koh and Percy Liang. 2017.
\newblock Understanding black-box predictions via influence functions.
\newblock In \emph{Proc. ICML}.

\bibitem[{Liu et~al.(2019)Liu, Ott, Goyal, Du, Joshi, Chen, Levy, Lewis,
  Zettlemoyer, and Stoyanov}]{liu2019roberta}
Yinhan Liu, Myle Ott, Naman Goyal, Jingfei Du, Mandar Joshi, Danqi Chen, Omer
  Levy, Mike Lewis, Luke Zettlemoyer, and Veselin Stoyanov. 2019.
\newblock Roberta: A robustly optimized bert pretraining approach.
\newblock \emph{arXiv preprint arXiv:1907.11692}.

\bibitem[{Min et~al.(2022)Min, Lyu, Holtzman, Artetxe, Lewis, Hajishirzi, and
  Zettlemoyer}]{min2022rethinking}
Sewon Min, Xinxi Lyu, Ari Holtzman, Mikel Artetxe, Mike Lewis, Hannaneh
  Hajishirzi, and Luke Zettlemoyer. 2022.
\newblock Rethinking the role of demonstrations: What makes in-context learning
  work?
\newblock In \emph{EMNLP}.

\bibitem[{Mostafazadeh et~al.(2017)Mostafazadeh, Roth, Louis, Chambers, and
  Allen}]{mostafazadeh2017lsdsem}
Nasrin Mostafazadeh, Michael Roth, Annie Louis, Nathanael Chambers, and James
  Allen. 2017.
\newblock Lsdsem 2017 shared task: The story cloze test.
\newblock In \emph{Proceedings of the 2nd Workshop on Linking Models of
  Lexical, Sentential and Discourse-level Semantics}, pages 46--51.

\bibitem[{Olsson et~al.(2022)Olsson, Elhage, Nanda, Joseph, DasSarma, Henighan,
  Mann, Askell, Bai, Chen et~al.}]{olsson2022context}
Catherine Olsson, Nelson Elhage, Neel Nanda, Nicholas Joseph, Nova DasSarma,
  Tom Henighan, Ben Mann, Amanda Askell, Yuntao Bai, Anna Chen, et~al. 2022.
\newblock In-context learning and induction heads.
\newblock \emph{arXiv preprint arXiv:2209.11895}.

\bibitem[{Ott et~al.(2021)Ott, Shleifer, Xu, Goyal, Duval, and Caggiano}]{FSDP}
Myle Ott, Sam Shleifer, Min Xu, Priya Goyal, Quentin Duval, and Vittorio
  Caggiano. 2021.
\newblock Fully sharded data parallel: faster ai training with fewer gpus.
\newblock \url{https://engineering.fb.com/2021/07/15/open-source/fsdp/}.

\bibitem[{Pan et~al.(2023)Pan, Gao, Chen, and Chen}]{Pan2023WhatIL}
Jane Pan, Tianyu Gao, Howard Chen, and Danqi Chen. 2023.
\newblock What in-context learning"learns"in-context: Disentangling task
  recognition and task learning.

\bibitem[{Petroni et~al.(2019)Petroni, Rockt{\"a}schel, Lewis, Bakhtin, Wu,
  Miller, and Riedel}]{Petroni2019LanguageMA}
Fabio Petroni, Tim Rockt{\"a}schel, Patrick Lewis, Anton Bakhtin, Yuxiang Wu,
  Alexander~H. Miller, and Sebastian Riedel. 2019.
\newblock Language models as knowledge bases?
\newblock In \emph{Proc. EMNLP}.

\bibitem[{Pillutla et~al.(2021)Pillutla, Swayamdipta, Zellers, Thickstun,
  Welleck, Choi, and Harchaoui}]{Pillutla2021MAUVEMT}
Krishna Pillutla, Swabha Swayamdipta, Rowan Zellers, John Thickstun, Sean
  Welleck, Yejin Choi, and Za{\"i}d Harchaoui. 2021.
\newblock Mauve: Measuring the gap between neural text and human text using
  divergence frontiers.
\newblock In \emph{Proc. NeurIPS}.

\bibitem[{Pruthi et~al.(2020)Pruthi, Liu, Sundararajan, and
  Kale}]{Pruthi2020EstimatingTD}
Garima Pruthi, Frederick Liu, Mukund Sundararajan, and Satyen Kale. 2020.
\newblock Estimating training data influence by tracking gradient descent.
\newblock In \emph{Proc. NeurIPS}.

\bibitem[{Razeghi et~al.(2022)Razeghi, IV, Gardner, and
  Singh}]{Razeghi2022ImpactOP}
Yasaman Razeghi, Robert L~Logan IV, Matt Gardner, and Sameer Singh. 2022.
\newblock Impact of pretraining term frequencies on few-shot reasoning.
\newblock \emph{ArXiv}, abs/2202.07206.

\bibitem[{Schick and Sch{\"u}tze(2021)}]{Schick2021ExploitingCF}
Timo Schick and Hinrich Sch{\"u}tze. 2021.
\newblock Exploiting cloze-questions for few-shot text classification and
  natural language inference.
\newblock In \emph{Proc. EACL}.

\bibitem[{Shin et~al.(2022)Shin, Lee, Ahn, Kim, Kim, Kim, Cho, Lee, Park, Ha
  et~al.}]{shin2022effect}
Seongjin Shin, Sang-Woo Lee, Hwijeen Ahn, Sungdong Kim, HyoungSeok Kim, Boseop
  Kim, Kyunghyun Cho, Gichang Lee, Woomyoung Park, Jung-Woo Ha, et~al. 2022.
\newblock On the effect of pretraining corpora on in-context learning by a
  large-scale language model.
\newblock \emph{arXiv preprint arXiv:2204.13509}.

\bibitem[{Shoeybi et~al.(2019)Shoeybi, Patwary, Puri, LeGresley, Casper, and
  Catanzaro}]{shoeybi2019megatron}
Mohammad Shoeybi, Mostofa Patwary, Raul Puri, Patrick LeGresley, Jared Casper,
  and Bryan Catanzaro. 2019.
\newblock Megatron-lm: Training multi-billion parameter language models using
  model parallelism.
\newblock \emph{arXiv preprint arXiv:1909.08053}.

\bibitem[{Socher et~al.(2013)Socher, Perelygin, Wu, Chuang, Manning, Ng, and
  Potts}]{socher2013recursive}
Richard Socher, Alex Perelygin, Jean Wu, Jason Chuang, Christopher~D Manning,
  Andrew~Y Ng, and Christopher Potts. 2013.
\newblock Recursive deep models for semantic compositionality over a sentiment
  treebank.
\newblock In \emph{Proceedings of the 2013 conference on empirical methods in
  natural language processing}, pages 1631--1642.

\bibitem[{Vaswani et~al.(2017)Vaswani, Shazeer, Parmar, Uszkoreit, Jones,
  Gomez, Kaiser, and Polosukhin}]{vaswani2017attention}
Ashish Vaswani, Noam Shazeer, Niki Parmar, Jakob Uszkoreit, Llion Jones,
  Aidan~N Gomez, {\L}ukasz Kaiser, and Illia Polosukhin. 2017.
\newblock Attention is all you need.
\newblock \emph{Advances in neural information processing systems}, 30.

\bibitem[{von Oswald et~al.(2022)von Oswald, Niklasson, Randazzo, Sacramento,
  Mordvintsev, Zhmoginov, and Vladymyrov}]{von2022transformers}
Johannes von Oswald, Eyvind Niklasson, Ettore Randazzo, Jo{\~a}o Sacramento,
  Alexander Mordvintsev, Andrey Zhmoginov, and Max Vladymyrov. 2022.
\newblock Transformers learn in-context by gradient descent.
\newblock \emph{arXiv preprint arXiv:2212.07677}.

\bibitem[{Wang et~al.(2022)Wang, Mishra, Alipoormolabashi, Kordi, Mirzaei,
  Arunkumar, Ashok, Dhanasekaran, Naik, Stap et~al.}]{wang2022benchmarking}
Yizhong Wang, Swaroop Mishra, Pegah Alipoormolabashi, Yeganeh Kordi, Amirreza
  Mirzaei, Anjana Arunkumar, Arjun Ashok, Arut~Selvan Dhanasekaran, Atharva
  Naik, David Stap, et~al. 2022.
\newblock Benchmarking generalization via in-context instructions on 1,600+
  language tasks.
\newblock \emph{arXiv preprint arXiv:2204.07705}.

\bibitem[{Wei et~al.(2023)Wei, Wei, Tay, Tran, Webson, Lu, Chen, Liu, Huang,
  Zhou, and Ma}]{Wei2023LargerLM}
Jerry~W. Wei, Jason Wei, Yi~Tay, Dustin Tran, Albert Webson, Yifeng Lu, Xinyun
  Chen, Hanxiao Liu, Da~Huang, Denny Zhou, and Tengyu Ma. 2023.
\newblock Larger language models do in-context learning differently.
\newblock \emph{ArXiv}, abs/2303.03846.

\bibitem[{Xie et~al.(2022)Xie, Raghunathan, Liang, and Ma}]{xie2022an}
Sang~Michael Xie, Aditi Raghunathan, Percy Liang, and Tengyu Ma. 2022.
\newblock \href {https://openreview.net/forum?id=RdJVFCHjUMI} {An explanation
  of in-context learning as implicit bayesian inference}.
\newblock In \emph{International Conference on Learning Representations}.

\bibitem[{Xiong et~al.(2019)Xiong, Wu, Wang, Kulkarni, Yu, Chang, Guo, and
  Wang}]{xiong2019tweetqa}
Wenhan Xiong, Jiawei Wu, Hong Wang, Vivek Kulkarni, Mo~Yu, Shiyu Chang,
  Xiaoxiao Guo, and William~Yang Wang. 2019.
\newblock Tweetqa: A social media focused question answering dataset.
\newblock \emph{arXiv preprint arXiv:1907.06292}.

\bibitem[{Zhang et~al.(2022{\natexlab{a}})Zhang, Zhang, Zhang, and
  Yang}]{zhang2022robustness}
Hongxin Zhang, Yanzhe Zhang, Ruiyi Zhang, and Diyi Yang. 2022{\natexlab{a}}.
\newblock Robustness of demonstration-based learning under limited data
  scenario.
\newblock \emph{arXiv preprint arXiv:2210.10693}.

\bibitem[{Zhang et~al.(2022{\natexlab{b}})Zhang, Roller, Goyal, Artetxe, Chen,
  Chen, Dewan, Diab, Li, Lin et~al.}]{zhang2022opt}
Susan Zhang, Stephen Roller, Naman Goyal, Mikel Artetxe, Moya Chen, Shuohui
  Chen, Christopher Dewan, Mona Diab, Xian Li, Xi~Victoria Lin, et~al.
  2022{\natexlab{b}}.
\newblock Opt: Open pre-trained transformer language models.
\newblock \emph{arXiv preprint arXiv:2205.01068}.

\bibitem[{Zhang et~al.(2015)Zhang, Zhao, and LeCun}]{Zhang2015CharacterlevelCN}
Xiang Zhang, Junbo~Jake Zhao, and Yann LeCun. 2015.
\newblock Character-level convolutional networks for text classification.
\newblock In \emph{Proc. NeurIPS}.

\end{thebibliography}
